\documentclass[11pt,a4paper]{article}
\usepackage[hyperref]{acl2019}
\usepackage{amsmath}
\usepackage{amssymb}
\usepackage{times}
\usepackage{latexsym}
\usepackage{url}
\usepackage{graphicx}
\usepackage{float}
\usepackage{caption}
\usepackage{subcaption}
\usepackage{booktabs}
\usepackage{adjustbox}
\usepackage{hyperref}

\aclfinalcopy % Uncomment this line for the final submission
\title{Cycle Text-to-Image GAN with BERT}

\author{
Trevor Tsue\\
\texttt{ttsue} \\
Computer Science Dept.
\And
Jason Li\\
\texttt{jasonkli}\\
Computer Science Dept.
\And
Samir Sen\\
\texttt{samirsen}\\
Computer Science Dept.
}

\date{9 June 2019}

\begin{document}
\maketitle
\begin{abstract}
We explore novel approaches to the task of image generation from their respective captions, building on state-of-the-art GAN architectures. Particularly, we baseline our models with the Attention-based GANs that learn attention mappings from words to image features. To better capture the features of the descriptions, we then built a novel cyclic design that learns an inverse function to maps the image back to original caption. Additionally, we incorporated recently developed BERT pretrained word embeddings as our initial text featurizer and observe a noticeable improvement in qualitative and quantitative performance compared to the Attention GAN baseline. \footnote{\href{https://github.com/suetAndTie/cycle-image-gan}{Cycle Image GAN Github}}
\end{abstract}

\section{Introduction}
\begin{figure*}
\centering
\begin{subfigure}{.5\textwidth}
  \centering
  \includegraphics[width=.75\linewidth]{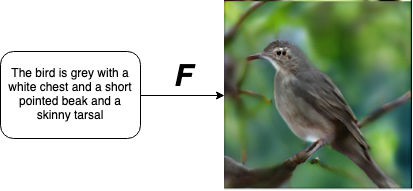}
  \caption{Text to Image}
  \label{fig:text2image}
\end{subfigure}%
\begin{subfigure}{.5\textwidth}
  \centering
  \includegraphics[width=.75\linewidth]{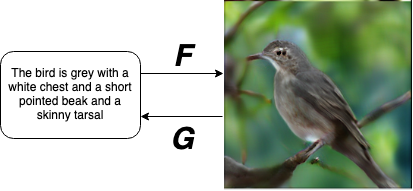}
  \caption{Text to Image to Text}
  \label{fig:text2image2text}
\end{subfigure}
\caption{Text to Image Generation Tasks}
\label{fig:test}
\end{figure*}

The goal of the text-to-image task is to generate realistic images given a text description. This problem has many possible applications ranging from computer-aided design to art generation \cite{DBLP:journals/corr/abs-1711-10485}.  Moreover, this multimodal problem is an interesting and important task in natural language understanding because it connects language to an understanding of the visual world.  

The problem can be naturally decomposed into two parts: embedding the text into a feature representation that captures relevant visual information and using that representation to generate an realistic image that corresponds to the text. This problem is particularly challenging for several reasons. For one, there is the issue of a domain gap between the text and image feature representations. Furthermore, there are a myriad of legitimate images that correspond to a single text description and part of the goal is to be able to capture the diversity in plausible images \cite{DBLP:journals/corr/ReedAYLSL16}.

Recent advances in the field of deep learning have made significant strides in this challenging task. In particular, recurrent architectures, such as LSTMs, can be used to learn feature representations from text and generative adversarial networks (GANs) can be used to create images conditioned on information. Nonetheless, the aforementioned challenges leave the text-to-image task an open problem. Adding to those problems, GANs, despite having widespread success in generative learning, often produce lower-resolution images, \cite{DBLP:journals/corr/ZhangXLZHWM16}, lack diversity in image generation, and fail to capture intricate details.

The goal of our work is to explore and compare state-of-the-art methods for addressing some of these problems.  These ideas include: stacking multiple GANs to sequentially learn higher resolution images; applying attention mechanisms to “focus” the generator on important parts of the text and to help bridge the domain gap; and unifying text-to-image and image-to-text in a single model. In addition to evaluating these ideas for diversity and quality of generated images, we also investigate the effect of using a pre-trained language model for contextualized word embeddings. Pretrained language models, such as BERT \cite{DBLP:journals/corr/abs-1810-04805} and ELMO \cite{DBLP:journals/corr/abs-1802-05365}, have revolutionized NLP as an effective means of transfer learning, similar to the impact of ImageNet on the field of computer vision.  As such, we sought to explore the potential benefit of these pretrained embeddings since most current approaches learn the word embeddings from scratch.

\section{Related Work}
Generative adversarial networks (GANs) are the most widely used model in generative learning. Originally proposed to generate realistic images, GANs consist of two neural networks, a generator and a discriminator, that effectively compete with one another in a zero-sum game. The discriminator attempts to distinguish real and fake images, while the generator tries to create images that fool the discriminator into classifying them as real \cite{NIPS2014_5423}. 

Numerous works have built off of the basic idea.  Some of these idea include the Conditional GANs, which pass a class label to both the generator and discriminator, unlike the original GAN where the generator creates an image solely from noise \cite{DBLP:journals/corr/MirzaO14}. GANs have also been used for style transfer between image domains. In this formulation, the generator is passed an image from a source domain and tries to fool the discriminator into thinking it is from the target domain. An extension to this idea is the CycleGAN, which learns to transfer from source to target and target back to source to ensure consistency and to stabilize training. This setup also ensures that important latent features are captured so that the source image can be reconstructed from the generated one \cite{originalcyclegan}. As we will see next, these ideas have a natural extension to the text-to-image problem.  

Reed et al. describe the first fully differentiable, end-to-end model that learns to construct images from text, building on conditional GANs \cite{DBLP:journals/corr/ReedAYLSL16}. They use a character-level convolutional-recurrent network to encode the input text. A fully-connected (FC) layer with a Leaky ReLU activation embeds the encoding into a lower dimension before concatenation with input noise drawn from a standard Gaussian, which is then fed into the generator. Instead of just training on pairs of either real images/matching text or fake images/matching text, they also train using pairs of real images with mismatched text.  This is to encourage the discriminator to not only generate realistic images regardless of the text, but also to create realistic images that match the text.  

Zhang et al. build upon this work in “StackGAN: Text to Photo-realistic Image Synthesis with Stacked Generative Adversarial Networks”. Inspired by other works that use multiple GANs for tasks such as scene generation, the authors used two stacked GANs for the text-to-image task \cite{DBLP:journals/corr/ZhangXLZHWM16}. The motivating intuition is that the Stage-I GAN produces a low-resolution outline of the desired image and the Stage-II GAN fills in the details of the sketch. In addition to the stacked architecture, they also propose a novel augmentation technique to address the aforementioned interpolation issues. Rather than using the text embedding directly (following a single FC layer) as done in Reed et al., they instead use an FC layer to produce a mean and variance before sampling from the normal distribution.  This sampling augments the data and increases the robustness. 

Another architecture developed by Xu et al. drew on the widespread success of attention-based models, particularly in NLP for tasks such as machine translation and image captioning. Their model, termed AttnGAN, introduces several novel ideas to the text-to-image task related to attention \cite{DBLP:journals/corr/abs-1711-10485}. Unlike previous approaches that focus on sentence-level encodings, AttnGAN extracts both sentence-level and word-level features from a bidirectional LSTM. In the stacked generator stages, multiplicative attention is performed over the encoded word vectors so the model can learn which words to attend to at each step. Finally, they add a Deep Attentional Multimodal Similarity Model, which is constructed to learn an attention-based matching score between the image-sentence pairs. 

Finally, Gorti et al. incorporate the ideas of stacking, attention, and cycle consistency in their state-of-the-art model, MirrorGAN \cite{mirrorgan}. Influenced by the CycleGAN architecture, the model adds an image-to-text component which acts as a sanity check that the image generated is indeed semantically consistent with the input caption text. The results demonstrate MirrorGAN's ability to train networks that can generate both higher quality images as well as image details which are semantically consistent with a provided caption (and in comparison with the true image for a given caption). 

MirrorGAN is the culmination of the work on the text-to-image problem. Nonetheless, there is room for improvement.   The works up until this point use word embeddings trained from scratch. With the advent of pretrained language models such as ELMO or BERT, a possible extension is to initialize the embeddings with deep, contextualized word vectors derived from BERT or ELMO.  In this work, we explore the effect of using BERT-derived word vectors.

\section{Data}
We used the 2011 Caltech-UCSD Birds 200 dataset (CUB-200), which contains 11,788 images of 200 different types of birds and is a widely used benchmark for text-to-image generation \cite{WahCUB_200_2011}. These images provide a boundary box and vary in size. Additionally, we have 10 text descriptions of the dataset downloaded from a github repository that serve as the text descriptions of the generated images \footnote{\href{https://github.com/taoxugit/AttnGAN}{taoxugit AttnGAN} \label{attngancode}}.

\section{Methods}
\subsection{Data Preprocessing}
We preprocess this data according the precedence set by StackGAN++ \cite{stackganpp}. This includes cropping all images to ensure all bounding boxes have at least a 0.75 object-image size ratio and then downsampled to 64x64, 128x128, 256x256. Then, the data is split into class disjoint train and test sets.

\subsection{Models}
\subsubsection{AttnGAN}
\begin{figure*}[ht]
    \centering
    \includegraphics[width=\textwidth]{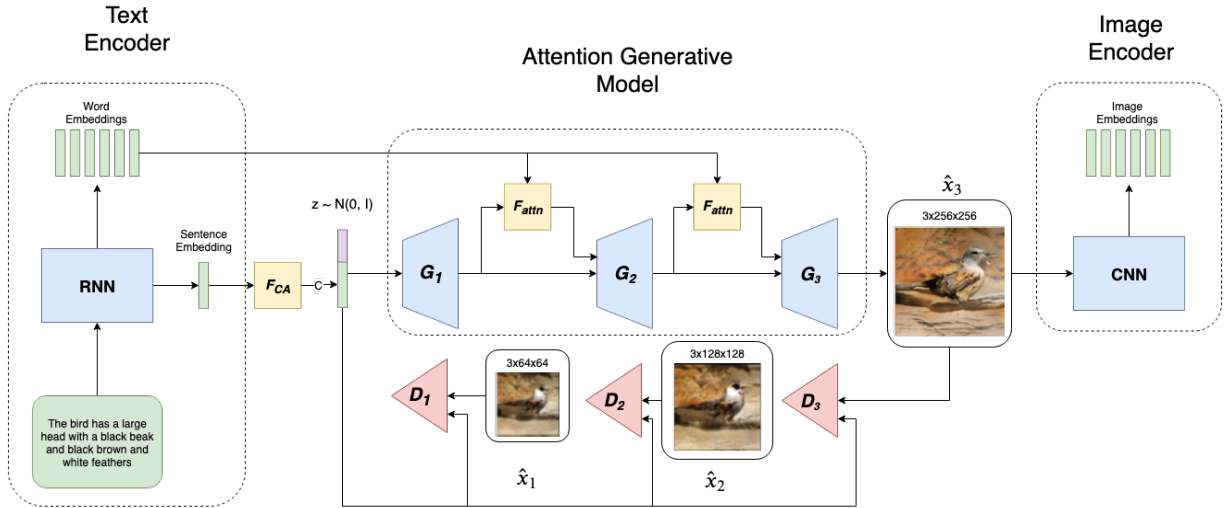}
    \caption{Attention GAN Architecture}
    \label{fig:attngan}
\end{figure*}
This first model combines both elements of the stack GAN \cite{DBLP:journals/corr/ZhangXLZHWM16} and attention \cite{DBLP:journals/corr/abs-1711-10485} . This attention GAN first embeds the caption and runs them through a LSTM, generating both word and sentence vectors. Using the Conditioning Augmentation first proposed in the StackGAN\cite{DBLP:journals/corr/SalimansGZCRC16}, we create a mean and variance from the sentence embedding via a fully-connected layer. We use this mean and variance to parameterize a normal distribution from which a sentence embedding sample is generated to pass into the GAN. This is used for regularization and to promote manifold smoothness. Additionally, we concatenate Gaussian noise to this new sentence embedding sample and pass into the generator.

With the StackGAN architecture, we stack three generators together, generating 64x64, 128x128, and 256x256, respectively. Additionally, for the second and third generator, we pass the image and the word embeddings through an attention module  to pass into the next generator \cite{DBLP:journals/corr/abs-1711-10485}. Each of these generators has a corresponding discriminator that take in both the original sentence embedding and the image. Finally, the 256x256 image is passed through an image encoder to generate local image features (a 17x17 feature map). These image features from the image encoder and word features from the text encoder combine to form the Deep Attentional Multimodal Similarity Model (DAMSM) and trained with an attention loss \cite{DBLP:journals/corr/abs-1711-10485}. For stability, we pretrained this DAMSM model.

\subsubsection{CycleGAN}
\begin{figure*}[ht]
    \centering
    \includegraphics[width=\textwidth]{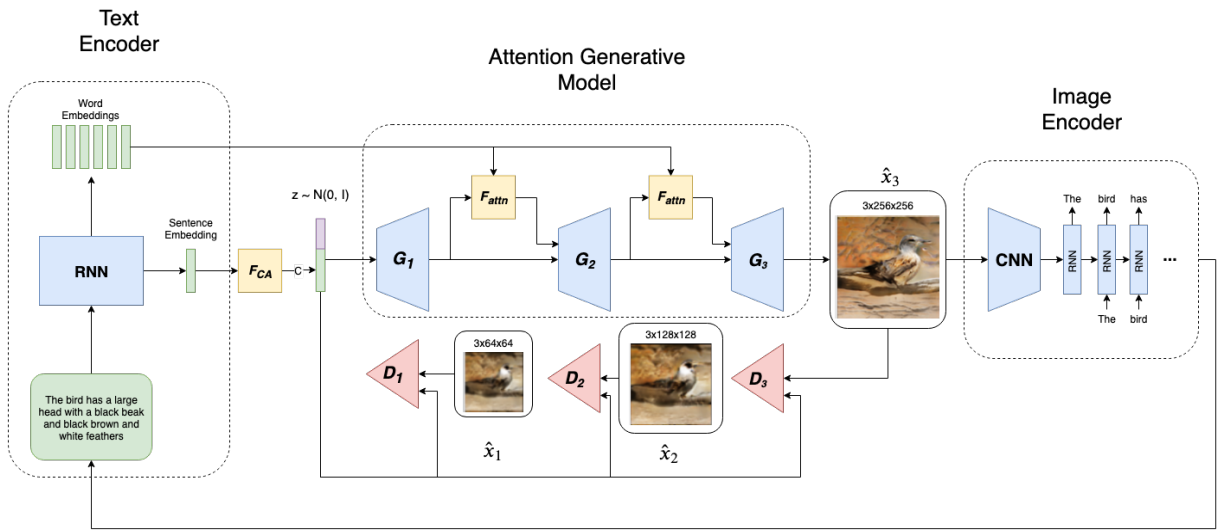}
    \caption{Cycle GAN Architecture}
    \label{fig:attngan}
\end{figure*}
Our CycleGAN combines the attention GAN and the original cycle GAN approach \cite{originalcyclegan}. By adding an RNN conditioned on the image features and the embedded captions, we attempt to return to text with the Semantic Text REgeneration and Alignment Module (STREAM) \cite{mirrorgan}. By learning this transition to the original text domain, we allow our images to better represent our captions, as they must hold the latent information to recreate the original caption \cite{mirrorgan} \footnote{\href{https://github.com/komiya-m/MirrorGAN}{komiya-m MirrorGAN}}. Additionally, we added the pretrained BERT encoding transformers, which we use instead of the standard word embeddings.

\subsubsection{BERT}
 Pretrained word vectors are a common component of many of NLP models. However, until recently, one primary limitation of these word vectors was that they only allowed for one context-independent embedding. One of the biggest game-changers in recent NLP research is the advent of deep, contextualized word vectors.  These vectors are derived from the internal states of deep, pretrained language models trained on massive corpuses of text and use the entire sequence to embed each word, not just the word itself.  A key premise to this idea was previous research that showed different layers of an LSTM language model captured different information, such as part of speech at the lower levels and context at the higher layers \cite{DBLP:journals/corr/abs-1802-05365}
 
Peters et al. were the first to introduce this idea with their language model, ELMO..  Their model was a deep, bidirectional LSTM with character-level convolutions.  This pretrained model could then be used for more specific tasks, where each word vector is computed as the (learned) weighted sum of the hidden states of the LSTM, using the entire input sequence as input. Then, tasks like sentiment classification could be done by simply adding a fully connected layer on top of ELMO \cite{DBLP:journals/corr/abs-1802-05365}.  Devlin et al. expanded on this work with the BERT model that replaces the bidirectional LSTM with a bidirectional Transformer \cite{DBLP:journals/corr/abs-1810-04805}.  The Transformer is another recent innovation in NLP that replaces the recurrent nature of LSTMs with positional encoding and blocks of self-attention, layer normalization, and fully connected layers \cite{DBLP:journals/corr/VaswaniSPUJGKP17}.  This architecture has become the de facto model for NLP, replacing LSTMs and standard RNNs in many cases.  The BERT model has been widely used to achieve state-of-the-art results in challenging tasks such as Question-Answer (QA).  In this work, we used a pretrained BERT model to obtain our embeddings and pass it through a fully connected layer before continuing in the CycleGAN architecture.

\subsection{Loss}
For a generator $G_i$ and the corresponding discriminator $D_i$, we have the following loss function which combines both a conditional and unconditional loss (conditioned on the sentence embedding):
\begin{align*}
\mathcal{L}_{G_i} = - \frac{1}{2} \mathbb{E}_{\hat x_i \sim p_{G_i}} [\log(D_i(\hat x_i))]\\
- \frac{1}{2} \mathbb{E}_{\hat x_i \sim p_{G_i}} [\log(D_i(\hat x_i, \bar e))]
\end{align*}
where $\hat x_i$ is the generated image and $\bar e$ is the sentence embedding.

Then, we have the following discriminator loss:
\begin{align*}
\mathcal{L}_{D_i} = - \frac{1}{2} \mathbb{E}_{x_i \sim p_{data_i}} [\log(D_i(x_i))]\\
- \frac{1}{2} \mathbb{E}_{\hat x_i \sim p_{G_i}} [\log(1 - D_i(\hat x_i))]\\
- \frac{1}{2} \mathbb{E}_{x_i \sim p_{data_i}} [\log(D_i(x_i))]\\
- \frac{1}{2} \mathbb{E}_{\hat x_i \sim p_{G_i, \bar e}} [\log(1 - D_i(\hat x_i, \bar e))]
\end{align*}

\subsubsection{AttnGAN Loss}
With the word embeddings matrix $e$ and the image embeddings $v$, we calculate a similarity score between the sentence: 
\begin{align}
s = e^\intercal v
\end{align}
We create a $s \in \mathbb{R}^{T \times 289}$ with $T$ as the number of words in the sentence and 289 referring to a flattened version of the 17x17 image feature map. We then normalize the similarity matrix
\begin{align}
\bar s_{ij} = \frac{\exp(s_{ij})}{\sum_{k=1}^T \exp(s_{kj})}
\end{align}
We build a context vector $c_i$, where that represent the image regions that relate to the $i$th word in the sentence.
\begin{align}
c_i = \sum_{j=1}^{289} \alpha_j v_j \text{, where } \alpha_j = \frac{\exp(\gamma \bar s_{ij})}{\sum_{k=1}^{289} \exp(\bar s_{ik})}
\end{align}
where $\gamma$ is a hyperparameter to pay attention to certain features in the regions. We then have an attention-driven image-text matching score matching the entire image $Q$ to the whole text description $D$ that utilizes the cosine similarity cosine$(c_i, e_i) = \frac{c_i^\intercal e_i}{\|c_i\| \|e_i\|}$
\begin{align}
R(Q,D) = \log \Big( \sum_{i=1}^T \exp(\gamma \text{cosine}(c_i, e_i))\Big)
\end{align}
We have the DAMSM probability between the different image-sentence pairs in the batch
\begin{align}
P(D_i|Q_i) = \frac{\exp(\gamma R(Q_i, D_i))}{\sum_{j=1}^M \exp(\gamma R(Q_i D_j))}\\
P(D_i|Q_i) = \frac{\exp(\gamma R(Q_i, D_i))}{\sum_{j=1}^M \exp(\gamma R(Q_j D_i))}
\end{align}
Therefore, our DAMSM combines the following
\begin{align}
\mathcal{L}_1^w = - \sum_{i=1}^M \log P(D_i | Q_i)\\
\mathcal{L}_2^w = - \sum_{i=1}^M \log P(Q_i | D_i)
\end{align}
We also define $\mathcal{L}_1^s$ and $\mathcal{L}_2^s$ the same as above but instead substituting $\bar e$ for $e$. Combining everything, we have the final loss of the attention generator
\begin{align}
\mathcal{L}_{DAMSM} = \mathcal{L}_1^w + \mathcal{L}_2^w + \mathcal{L}_1^s + \mathcal{L}_2^s
\end{align}
\begin{align}
\mathcal{L} = \mathcal{L}_G + \lambda \mathcal{L}_{DAMSM} \text{, where } \mathcal{L}_G = \sum_{i=1}^3 \mathcal{L}_{G_i}
\end{align}

\subsubsection{CycleGAN Loss}
In addition to the loss of the AttnGAN, we add an additional cross entropy loss to correctly predict the output word in the caption recreation
\begin{align*}
\mathcal{L}_{CE} = - \frac{1}{M} \sum_{i=1}^M \sum_{c=1}^{|V|} y_c^{(i)} \log (\hat y_c^{(i)})\\
\mathcal{L} = \mathcal{L}_G + \lambda \mathcal{L}_{DAMSM} + \lambda \mathcal{L}_{CE}
\end{align*}
where $M$ represents the batch size, $|V|$ is the size of the vocab, $y_c^{(i)}$ is the binary label of the $c$-th class of the $i$-th example, and $\hat y_c^{(i)}$ is the model probability output of the $c$-th class of the $i$-th example.

\subsection{Evaluation Metrics}
\subsubsection{Inception Score}
To assess our models, we use the Inception score, which is a widely used ad hoc metric for generative models \cite{DBLP:journals/corr/SalimansGZCRC16}. The Inception score uses a pretrained Inception model that is fine-tuned to the specific dataset being used. The Inception score is computed by exponentiating the KL-divergence between the conditional distribution p(y $|$ x) and marginal distribution p(y), where y is the class label predicted by the Inception model and x is a generated sample. The intuition is that a good generative model should produce images with a  conditional label distribution that has low entropy relative to the marginal distribution. In other words, we want images that can be easily classified into a category by the model but also create images that belong to many different classes. 
\begin{align*}
D_{KL} (P \| Q) = - \sum_{x \in \mathcal{X}} P(x) \log \Big(\frac{Q(x)}{P(x)}\Big)\\
IS(G) = \exp \Big(\mathbb{E}_{\bold x \sim p_G} D_{KL} (p(y|\bold x) \| p(y)) \Big)
\end{align*}

The score rewards images that have greater variety and has been shown to be well-correlated with human evaluations of realistic quality. We randomly select 20 captions for each class and use our trained model to generate images, which is then fed into the Inception model to generate the distributions and to compute the score.

\subsubsection{Mean Opinion Score (MOS)}
Nonetheless, the Inception score cannot capture how well the generated images reflect accurate conditioning on the input text. Thus, we have humans examine the perceptual quality of images  as well as their correspondence to the input task with the Mean Opinion Score \cite{DBLP:journals/corr/LedigTHCATTWS16}. Specifically, we asked $n=10$ subjects to rate the quality of images on a scale from 1 (poor quality) to 5 (high quality). We showed them 20 images from the ground truth, 20 from the AttnGAN, and 20 from the CycleGAN, along with corresponding captions, in random order, and averaged them to report the MOS.

\section{Results}
\begin{figure}[ht]
\includegraphics[width=.55\textwidth]{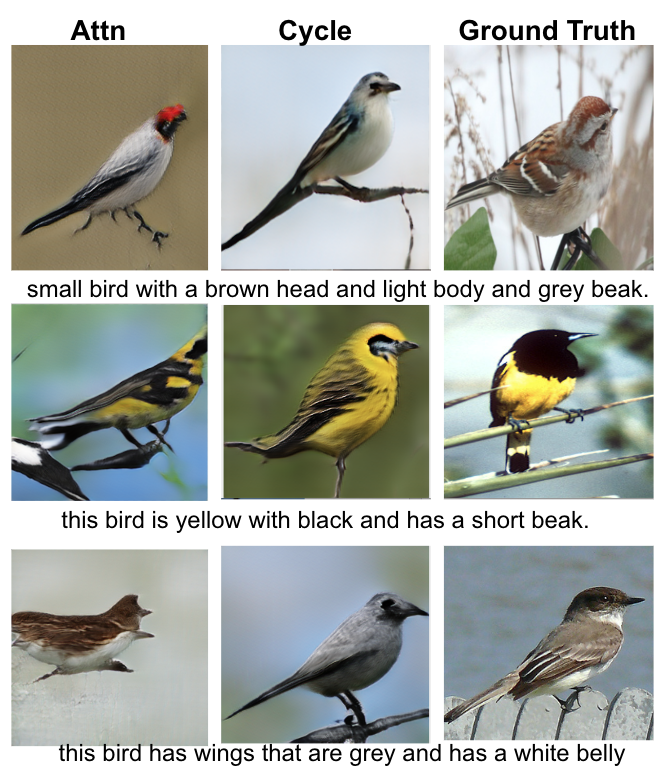}
\caption{Generated Images from the models}
\label{fig:generated}
\end{figure}

We trained the AttnGAN model over 100 epochs using Adam optimization to train the generator and all the discriminators. CycleGAN was trained over 100 epochs using the same generator and discriminator optimizers with betas of 0.5 and 0.999. For the AttnGAN, we pretrained the DAMSM architecture for 200 epochs. For the CycleGAN, we pretrained the STREAM architecture for 100 epochs. We used pretrained BERT embeddings only for the CycleGAN implementation, while initialized randomly initialized embeddings which were trained in the AttnGAN. \\

We report both the inception v3 scores, computed as a average measure of divergence with the true distribution of bird images of the generated test outputs for each of the models as well as the qualitative MOS scores from peer judges, reported below. We see that the CycleGAN trained with BERT embeddings had the strongest performance overall across the proposed metrics, and display generated samples from our model along with their representative ground truth image labels. \\

\begin{figure}[ht]
\centering
\begin{adjustbox}{width=.48\textwidth}
\begin{tabular}{cccc}
\toprule
&   \multicolumn{3}{c}{\textbf{Inception Score}}\\ \cmidrule{2-4}
\textbf{Model}  &   Epoch 0 & Epoch 50  & Epoch 100\\ \midrule
Ground Truth    &   11.63 &  - & - \\
AttnGAN         &   0.94    &   2.78    &   3.92    \\  
CycleGAN w/ BERT    &   1.05   &   5.48    &   5.92 \\
\bottomrule
\end{tabular}
\end{adjustbox}
\caption{Inception Scores of Models}
\label{fig:inceptionscores}
\end{figure}

\begin{figure}[ht]
\centering
\begin{adjustbox}{width=0.48\textwidth}
\begin{tabular}{cc}
\toprule
\textbf{Model}  & \textbf{MOS (n=10)}\\ \midrule
Ground Truth    & 4.7\\
AttnGAN         & 3.6\\
CycleGAN w/ BERT & 3.9\\
\bottomrule
\end{tabular}
\end{adjustbox}
\caption{Mean Opinion Score (MOS) of Models with $n=10$ subjects}
\label{fig:mos}
\end{figure}

We save model weights for the CycleGAN model with BERT text features every 25 epochs and compute Inception scores on the validation set during training. In Figure 7, we observe the CycleGAN inception score leveling, but still increasing as we approach 100 epochs of training. 

\begin{figure}[ht]
\centering
\includegraphics[width=.50\textwidth]{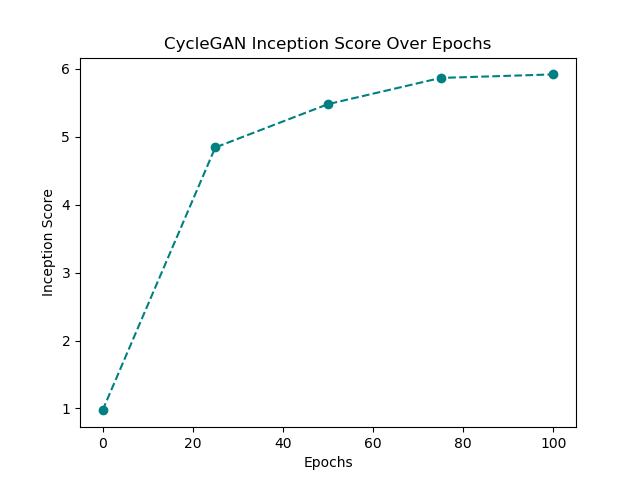}
\caption{CycleGAN Inception Scores}
\label{fig:inception_score}
\end{figure}

\section{Discussion}

Examining several images output from the AttnGAN and CycleGAN (with BERT) in Figure 4, we can see some clear improvements from AttnGAN to CycleGAN.  For one, the CycleGAN model generally produces clearer, more realistic looking images relative to AttnGAN. Furthermore, the CycleGAN model appears to be more precise with respect to details. We can see that for the AttnGAN model, the colors are occasionally incorrect (presence of red in the top image and brown instead of grey in the bottom). Additionally, AttnGAN images lack the level of detail in the beak present in the CycleGAN model. 

We found the CycleGAN with pretrained BERT embeddings was able to outperform AttnGAN on the test set in both the Inception score (Figure 5) and in the Mean Opinion Score (Figure 6). In particular, with respect to Inception scores, we found that CycleGAN with BERT was able to reach a higher score and train significantly faster, as indicated by the scores at 50 epochs (5.48 vs 2.78) and at 100 epochs (5.92 vs 3.92).  In general, a higher Inception score reflects greater variety as well as distinctly capturing unique features, but we note that an ideal quantitative metric remains elusive for this task, particularly in capturing the correspondence between the image and caption.  The higher performance of the CycleGAN with BERT on qualitative, human evaluation indicates some level of improvement in the image-text correspondence. One limitation of our work is that we were not able to train until convergence of the Inception score for comparison in the limit, which we note as a possible avenue for future work.

\section{Conclusion}
In this paper, we investigate the text-to-image generation task by experimenting with state-of-the-art architectures and incorporating the latest innovations in NLP, namely, the use of deep contextualized word vectors from pretrained languagew models, such as BERT. Our baseline model is the AttnGAN, which utilizes several key features, including stacking of GANs to progressively learn more detail at higher resolution and attention over word features, a technique that has found widespread success in a variety of NLP tasks.  Our main model adds two additional features: a cyclic architectures that adds the image-to-text task in addition to the test-to-image task, and the use of contexualized word embeddings from a pretrained BERT. Through both qualitative and quantitative metrics, we found that the addition of these features showed improved generation of images conditioned on the text and had faster learning. 

For future work, it would be useful to train the models longer until convergence in some metric (such as Inception score) is reached for complete analysis. In addition, we did not do any hyperparameter tuning due to time constraints and thus further improvement may be found through a hyperparameter search. Further, with more time, it would also be interesting to perform ablation studies on our full model to show the additional gain, if any, achieved from adding only BERT or only the cyclic architecture to AttnGAN.

\section{Acknowledgements}
We would like to thank the instructors and TAs for designing and running the course. In particular we would like to thank Ignacio Cases for guiding us on this project. 

\section{Authorship Statement}
Trevor Tsue: Coded the AttnGAN and CycleGAN, trained models, made architecture diagrams, wrote architecture and loss.\\\\
Jason Li:  Performed literature review, proposed and coded/integrated BERT extension with CycleGAN; wrote intro, related work, discussion, conclusion\\\\
Samir Sen: AttnGAN implementation and training. Worked on incorporating BERT encodings within the attention GAN text featurization and built inception network for capturing key metrics across models. Data cleaning and visualization. Results, discussion, abstract. \\

\bibliography{acl2019}

\begin{thebibliography}{14}
\expandafter\ifx\csname natexlab\endcsname\relax\def\natexlab#1{#1}\fi

\bibitem[{Devlin et~al.(2018)Devlin, Chang, Lee, and
  Toutanova}]{DBLP:journals/corr/abs-1810-04805}
Jacob Devlin, Ming{-}Wei Chang, Kenton Lee, and Kristina Toutanova. 2018.
\newblock \href {http://arxiv.org/abs/1810.04805} {{BERT:} pre-training of deep
  bidirectional transformers for language understanding}.
\newblock \emph{CoRR}, abs/1810.04805.

\bibitem[{Goodfellow et~al.(2014)Goodfellow, Pouget-Abadie, Mirza, Xu,
  Warde-Farley, Ozair, Courville, and Bengio}]{NIPS2014_5423}
Ian Goodfellow, Jean Pouget-Abadie, Mehdi Mirza, Bing Xu, David Warde-Farley,
  Sherjil Ozair, Aaron Courville, and Yoshua Bengio. 2014.
\newblock \href
  {http://papers.nips.cc/paper/5423-generative-adversarial-nets.pdf}
  {Generative adversarial nets}.
\newblock In Z.~Ghahramani, M.~Welling, C.~Cortes, N.~D. Lawrence, and K.~Q.
  Weinberger, editors, \emph{Advances in Neural Information Processing Systems
  27}, pages 2672--2680. Curran Associates, Inc.

\bibitem[{Ledig et~al.(2016)Ledig, Theis, Huszar, Caballero, Aitken, Tejani,
  Totz, Wang, and Shi}]{DBLP:journals/corr/LedigTHCATTWS16}
Christian Ledig, Lucas Theis, Ferenc Huszar, Jose Caballero, Andrew~P. Aitken,
  Alykhan Tejani, Johannes Totz, Zehan Wang, and Wenzhe Shi. 2016.
\newblock \href {http://arxiv.org/abs/1609.04802} {Photo-realistic single image
  super-resolution using a generative adversarial network}.
\newblock \emph{CoRR}, abs/1609.04802.

\bibitem[{Mirza and Osindero(2014)}]{DBLP:journals/corr/MirzaO14}
Mehdi Mirza and Simon Osindero. 2014.
\newblock \href {http://arxiv.org/abs/1411.1784} {Conditional generative
  adversarial nets}.
\newblock \emph{CoRR}, abs/1411.1784.

\bibitem[{Peters et~al.(2018)Peters, Neumann, Iyyer, Gardner, Clark, Lee, and
  Zettlemoyer}]{DBLP:journals/corr/abs-1802-05365}
Matthew~E. Peters, Mark Neumann, Mohit Iyyer, Matt Gardner, Christopher Clark,
  Kenton Lee, and Luke Zettlemoyer. 2018.
\newblock \href {http://arxiv.org/abs/1802.05365} {Deep contextualized word
  representations}.
\newblock \emph{CoRR}, abs/1802.05365.

\bibitem[{Qiao et~al.(2019)Qiao, Zhang, Xu, and Tao}]{mirrorgan}
Tingting Qiao, Jing Zhang, Duanqing Xu, and Dacheng Tao. 2019.
\newblock \href {http://arxiv.org/abs/1903.05854} {Mirrorgan: Learning
  text-to-image generation by redescription}.
\newblock \emph{CoRR}, abs/1903.05854.

\bibitem[{Reed et~al.(2016)Reed, Akata, Yan, Logeswaran, Schiele, and
  Lee}]{DBLP:journals/corr/ReedAYLSL16}
Scott~E. Reed, Zeynep Akata, Xinchen Yan, Lajanugen Logeswaran, Bernt Schiele,
  and Honglak Lee. 2016.
\newblock \href {http://arxiv.org/abs/1605.05396} {Generative adversarial text
  to image synthesis}.
\newblock \emph{CoRR}, abs/1605.05396.

\bibitem[{Salimans et~al.(2016)Salimans, Goodfellow, Zaremba, Cheung, Radford,
  and Chen}]{DBLP:journals/corr/SalimansGZCRC16}
Tim Salimans, Ian~J. Goodfellow, Wojciech Zaremba, Vicki Cheung, Alec Radford,
  and Xi~Chen. 2016.
\newblock \href {http://arxiv.org/abs/1606.03498} {Improved techniques for
  training gans}.
\newblock \emph{CoRR}, abs/1606.03498.

\bibitem[{Vaswani et~al.(2017)Vaswani, Shazeer, Parmar, Uszkoreit, Jones,
  Gomez, Kaiser, and Polosukhin}]{DBLP:journals/corr/VaswaniSPUJGKP17}
Ashish Vaswani, Noam Shazeer, Niki Parmar, Jakob Uszkoreit, Llion Jones,
  Aidan~N. Gomez, Lukasz Kaiser, and Illia Polosukhin. 2017.
\newblock \href {http://arxiv.org/abs/1706.03762} {Attention is all you need}.
\newblock \emph{CoRR}, abs/1706.03762.

\bibitem[{Wah et~al.(2011)Wah, Branson, Welinder, Perona, and
  Belongie}]{WahCUB_200_2011}
C.~Wah, S.~Branson, P.~Welinder, P.~Perona, and S.~Belongie. 2011.
\newblock {The Caltech-UCSD Birds-200-2011 Dataset}.
\newblock Technical Report CNS-TR-2011-001, California Institute of Technology.

\bibitem[{Xu et~al.(2017)Xu, Zhang, Huang, Zhang, Gan, Huang, and
  He}]{DBLP:journals/corr/abs-1711-10485}
Tao Xu, Pengchuan Zhang, Qiuyuan Huang, Han Zhang, Zhe Gan, Xiaolei Huang, and
  Xiaodong He. 2017.
\newblock \href {http://arxiv.org/abs/1711.10485} {Attngan: Fine-grained text
  to image generation with attentional generative adversarial networks}.
\newblock \emph{CoRR}, abs/1711.10485.

\bibitem[{Zhang et~al.(2016)Zhang, Xu, Li, Zhang, Huang, Wang, and
  Metaxas}]{DBLP:journals/corr/ZhangXLZHWM16}
Han Zhang, Tao Xu, Hongsheng Li, Shaoting Zhang, Xiaolei Huang, Xiaogang Wang,
  and Dimitris~N. Metaxas. 2016.
\newblock \href {http://arxiv.org/abs/1612.03242} {Stackgan: Text to
  photo-realistic image synthesis with stacked generative adversarial
  networks}.
\newblock \emph{CoRR}, abs/1612.03242.

\bibitem[{Zhang et~al.(2017)Zhang, Xu, Li, Zhang, Wang, Huang, and
  Metaxas}]{stackganpp}
Han Zhang, Tao Xu, Hongsheng Li, Shaoting Zhang, Xiaogang Wang, Xiaolei Huang,
  and Dimitris~N. Metaxas. 2017.
\newblock \href {http://arxiv.org/abs/1710.10916} {Stackgan++: Realistic image
  synthesis with stacked generative adversarial networks}.
\newblock \emph{CoRR}, abs/1710.10916.

\bibitem[{Zhu et~al.(2017)Zhu, Park, Isola, and Efros}]{originalcyclegan}
Jun{-}Yan Zhu, Taesung Park, Phillip Isola, and Alexei~A. Efros. 2017.
\newblock \href {http://arxiv.org/abs/1703.10593} {Unpaired image-to-image
  translation using cycle-consistent adversarial networks}.
\newblock \emph{CoRR}, abs/1703.10593.

\end{thebibliography}
\bibliographystyle{acl_natbib}

\end{document}